\documentclass[sigconf]{acmart}

\copyrightyear{2021} 
\acmYear{2021} 
\setcopyright{acmcopyright}\acmConference[ICMI '21]{Proceedings of the 2021 International Conference on Multimodal Interaction}{October 18--22, 2021}{Montréal, QC, Canada}
\acmBooktitle{Proceedings of the 2021 International Conference on Multimodal Interaction (ICMI '21), October 18--22, 2021, Montréal, QC, Canada}
\acmPrice{15.00}
\acmDOI{10.1145/3462244.3479892}
\acmISBN{978-1-4503-8481-0/21/10}

\usepackage{multirow}
\newcommand{\tabincell}[2]{\begin{tabular}{@{}#1@{}}#2\end{tabular}}

\begin{document}

%%
%% The "title" command has an optional parameter,
%% allowing the author to define a "short title" to be used in page headers.
\title{Efficient Deep Feature Calibration for Cross-Modal Joint Embedding Learning}

%%
%% The "author" command and its associated commands are used to define
%% the authors and their affiliations.
%% Of note is the shared affiliation of the first two authors, and the
%% "authornote" and "authornotemark" commands
%% used to denote shared contribution to the research.

\author{Zhongwei Xie$^{+,*}$, Ling Liu$^{+}$, Lin Li$^{*}$, Luo Zhong$^{*}$}
\affiliation{%
  \institution{$^{+}$School of Computer Science, Georgia Institute of Technology, Atlanta, Georgia, USA \\
$^{*}$School of Computer Science and Technology, Wuhan University of Technology, Wuhan, Hubei, China}
  \country{}}
  \email{zhongweixie@gatech.edu, lingliu@cc.gatech.edu, {cathylilin, zhongluo}@whut.edu.cn}

%%
%% By default, the full list of authors will be used in the page
%% headers. Often, this list is too long, and will overlap
%% other information printed in the page headers. This command allows
%% the author to define a more concise list
%% of authors' names for this purpose.
\renewcommand{\shortauthors}{Xie and Liu, et al.}

%%
%% The abstract is a short summary of the work to be presented in the
%% article.
\begin{abstract}
This paper introduces a two-phase deep feature calibration framework for efficient learning of semantics enhanced text-image cross-modal joint embedding, which clearly separates the deep feature calibration in data preprocessing from training the joint embedding model. We use the Recipe1M dataset for the technical description and empirical validation. In preprocessing, we perform deep feature calibration by combining deep feature engineering with semantic context features derived from raw text-image input data. We leverage LSTM to identify key terms, NLP methods to produce ranking scores for key terms before generating the key term feature. We leverage wideResNet50 to extract and encode the image category semantics to help semantic alignment of the learned recipe and image embeddings in the joint latent space. In joint embedding learning, we perform deep feature calibration by optimizing the batch-hard triplet loss function with soft-margin and double negative sampling, also utilizing the category-based alignment loss and discriminator-based alignment loss. Extensive experiments demonstrate that our SEJE approach with the deep feature calibration significantly outperforms the state-of-the-art approaches.
\end{abstract}

%%
%% The code below is generated by the tool at http://dl.acm.org/ccs.cfm.
%% Please copy and paste the code instead of the example below.
%%


%%
%% Keywords. The author(s) should pick words that accurately describe
%% the work being presented. Separate the keywords with commas.

%% A "teaser" image appears between the author and affiliation
%% information and the body of the document, and typically spans the
%% page.

%%
%% This command processes the author and affiliation and title
%% information and builds the first part of the formatted document.
\maketitle

\begin{figure*}[t] 
  \centering 
  \includegraphics[scale=0.22]{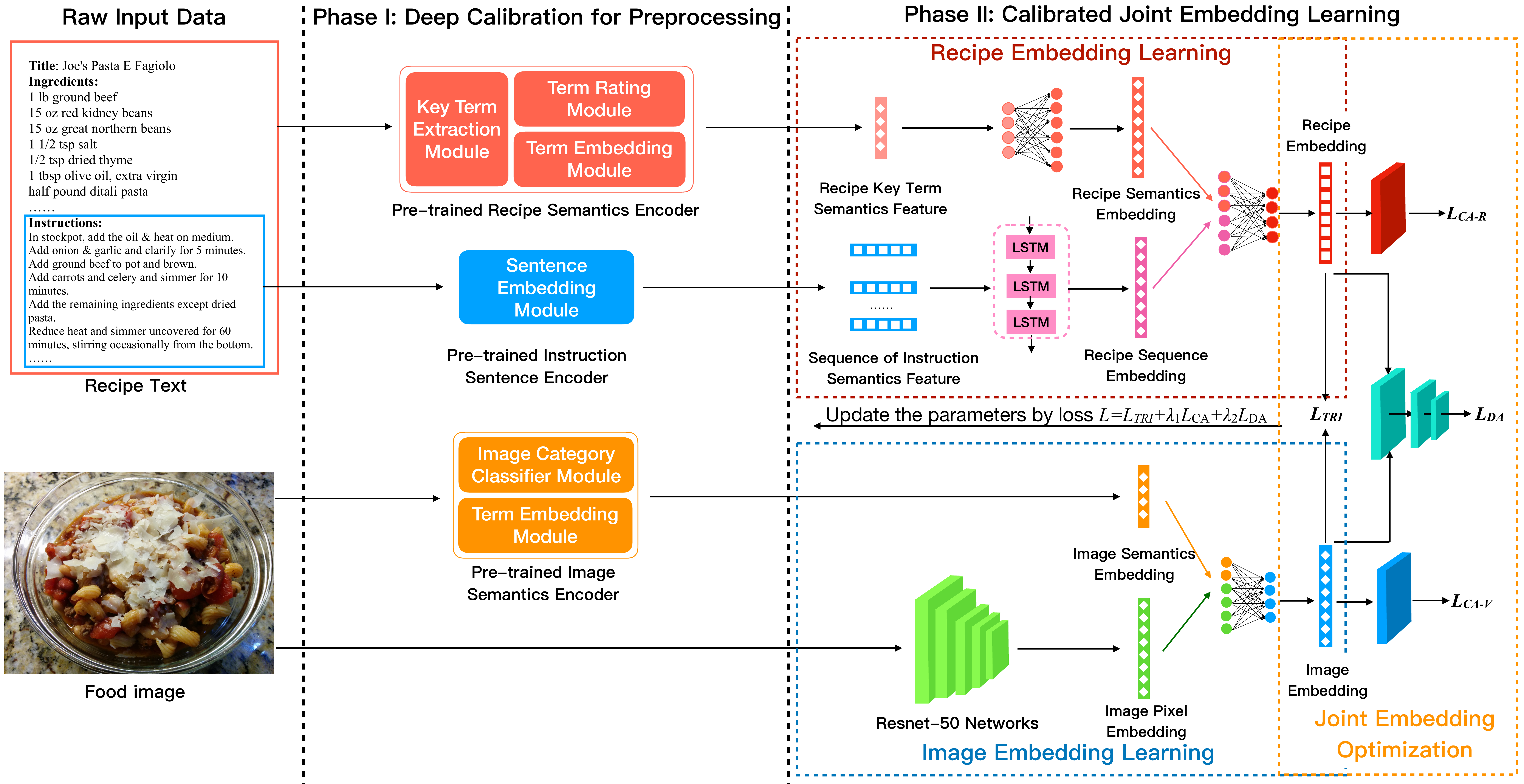}  
  \caption{The framework of SEJE cross-modal embedding learning approach.}
  \label{general_framework} 
\end{figure*}

\section{Introduction}

The cross-modal embedding learning problem belongs to the family of unsupervised learning~\cite{Yan+CVPR2015} and it aims to train a multi-modal joint embedding model over an unlabeled multi-modal dataset, such as Recipe1M~\cite{Salvador+CVPR2017_JESR}, by mapping features of different modalities onto the same latent space for similarity-based assessment, such as cross-modal retrieval. Most recipes provide ingredients with their quantities and cooking instructions on how ingredients are prepared and cooked (e.g., steamed or deep-fried), providing a new source of references for food intake tracking and health management. Among the few proposals to address the cross-modal embedding problem using Recipe1M, most of the recent approaches utilize the same embedding learning process to obtain the recipe and image embedding as that of~\cite{Salvador+CVPR2017_JESR}, while differ mainly in similarity metric learning to regulate the joint embeddings for the cooking recipe and food image. However, few existing approaches have utilized the feature engineering techniques to incorporate the additional semantics in the input recipe and image data into the modality-specific embedding learning and joint embedding loss optimization.
 
In this paper, we argue that calibrating the modality-specific embeddings by leveraging the additional semantics from the input data can effectively optimize the modality-specific embedding learning and regulate the cross-modal similarity loss function during iterative learning of the cross-modal joint embedding. 
We present a novel two-phase deep feature calibration framework for efficient learning of Semantics Enhanced text-image cross-modal Joint Embedding, coined as SEJE. This paper makes three main contributions. 
{\em First}, in data preprocessing, we introduce the LSTM-based key term extraction module, term rating module and term embedding module to identify those key terms that can uniquely distinguish its recipe from other recipes and utilize the sentence embedding module to generate a vector representation for each sentence in the cooking instruction of the input recipe, prior to the LSTM-based recipe embedding learning process.   
{\em Second}, we also develop the image category classifier module and term embedding module to incorporate the additional semantics into the image embedding, and iteratively align the image embedding semantically closer to the associated recipe embedding in the learned joint latent space.
{\em Third}, we propose the deep feature calibrated loss optimizations to regulate the joint embedding learning, comprising an improved batch-hard triplet loss empowered with soft-margin function and double negative sampling strategy,  category-based alignment loss and discriminator-based alignment loss, which boost the efficiency and performance of cross-modality joint embedding learning. 
Extensive experiments are conducted for cross-modal retrieval tasks on the Recipe1M benchmark dataset. The evaluation results show that empowered by the two-phase deep feature calibration techniques, our SEJE approach outperforms existing representative methods in terms of both image-to-recipe and recipe-to-image cross-modal retrieval performance.

\section{Related Work}  
The first large-scale corpus of structured recipe dataset is introduced in JESR~\cite{Salvador+CVPR2017_JESR}, which proposes to learn a joint embedding by combining a pairwise cosine loss with a semantic regularization constraint. A few follow-up efforts improve the result of JESR by upgrading the cosine similarity between the recipe text embedding and the image embedding, while adopting the same LSTM networks and CNN networks to generate the recipe embedding and image embedding respectively. AMSR~\cite{JinJinChen+MM2018_AMSR} improves JESR by combining hierarchical attention on the recipes with a simple triplet loss for joint embedding. AdaMine~\cite{Carvalho+SIGIR2018_AdaMine} optimizes the simple (batch-all) triplet loss with a class-level triplet loss and zero triplets removal strategy.
ACME~\cite{Hao+CVPR2019_ACME} further improves AdaMine and JESR by extending the joint optimization with adversarial cross-modal embedding \cite{Wang-2017} and a hard sample mining strategy. Recipe Retrieval with visual query of ingredients (Img2img+JESR)~\cite{lien2020recipe}, Recipe Retrieval with GAN (R$^2$GAN)~\cite{zhu2019r2gan} and Modality-Consistent Embedding Network (MCEN)~\cite{fu2020mcen} are recent additions, although they cannot outperform ACME and AdaMine.

 \begin{figure*}[tbp]
  \centering
  \includegraphics[scale=0.42]{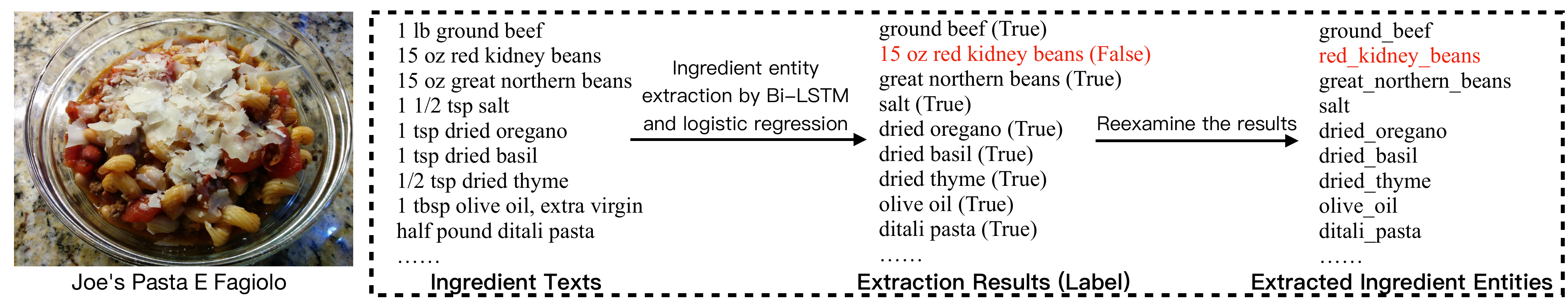} 
  \caption{{ The workflow of extracting the ingredient entities is depicted in the dashed box for the given recipe and the matched food image is listed on the left. }} 
  \label{miss_ingr} 
\end{figure*} 

\section{Semantics Enhanced Joint Embedding}

Let $\mathbf{R}$ and $\mathbf{V}$ denote the recipe domain and image domain respectively. For a set $T$ of recipe-image pairs ($\mathbf{r}_{i}$, $\mathbf{v}_i$), where a recipe $\mathbf{r}_{i}\in \mathbf{R}$ and an image $\mathbf{v}_{i}\in \mathbf{V}$ ($1\leq i\leq T$), we want to jointly learn two embedding functions, $\mathbf{E}_{V}: \mathbf{V}\rightarrow \mathbb{R}^{d}$ and $\mathbf{E}_{R}: \mathbf{R}\rightarrow \mathbb{R}^{d}$, which encode each pair of raw recipe and food image into two $d$-dimensional vectors in the latent representation space $\mathbb{R}^{d}$.
The two embedding functions should satisfy the following condition: For $1 \leq i, j\leq T$, the distance between a recipe $\mathbf{r}_{i}$ and an image $\mathbf{v}_j$ in the latent $d$-dimension embedding space should be closer when $i=j$ and more distant when $i\neq j$. 

We argue that in order to ensure efficient learning of the two embedding functions:  $\mathbf{E}_{R}$ and $\mathbf{E}_{V}$, we need to perform both deep feature calibrations in the data preprocessing phase (Phase I) and joint embedding learning phase (Phase II). In Phase I of SEJE, we combine LSTM, deep NLP models, wideResNet50 and word2vec to extract key term semantics and sequence semantics from recipe title, ingredients, cooking instructions, and image category semantics from associated food images. In Phase II of SEJE, we leverage the semantic features extracted from Phase I for dual purposes: (1) to enhance the quality of each modality-specific embedding to more accurately capture the semantic association of input recipe-image pair; and (2) to improve the joint embedding loss optimization. Figure~\ref{general_framework} gives a sketch of our SEJE framework. 

In data preprocessing, we introduce the LSTM-based key term extraction module, the term rating module and the term embedding module to preprocess the training data in three steps: (i) identifying the discriminative key terms in the recipe text, which can uniquely distinguish this recipe from other recipes in the million recipes dataset,  (ii) generating a vector representation for each key-term (such as key ingredients in a recipe), and generating a vector representation for each sentence in cooking instruction of the input recipe; and (iii) generating a vector representation of the semantic category or caption description for each image in the input data. The vectors for recipe text will be fed to the text embedding module and the vectors for the image in the same recipe will be fed together with the image pixel vector representation into the image embedding module during joint embedding training. Then deep feature calibrated loss optimization will be applied.

\subsection{Deep Calibration for Preprocessing}
%% Semantic Feature Engineering} 
In this work, we mainly extract and leverage three semantic features from the raw input data: recipe instruction semantics, key term semantics and image category semantics.

\subsubsection{Recipe Instruction Semantics} 
Considering each recipe has a list of cooking instructions that are quite lengthy, using a single LSTM is not appropriate to obtain the overall instruction embedding since the gradients might be diminished over so many time steps. Therefore, we first obtain the word-level semantics feature of each instruction sentence by a pre-trained instruction sentence embedding model, which is a product of the auto-encoder model built upon LSTM and skip-thoughts~\cite{Kiro-2015} that encodes a sentence and uses its encoding as context when predicting its previous and next sentences. Then the obtained sequence of each instruction sentence semantics feature is fed into the recipe embedding learning process.

\subsubsection{Recipe Key Term Semantics} 
The recipe key term semantics feature is utilized to capture the level of discrimination significance contributed by each key term to its recipe text. The inherent problem of LSTM-based existing works is that they learn every word in a recipe similarly, overlooking the fact that some descriptions of words are not visually observable w.r.t. the corresponding food images, or not discriminative with respect to either the recipe (e.g., sugar, salt) or the way of being used in the cooking instruction. The recipe semantics embedding can be leveraged to correct errors made in sequence feature engineering, which are obtained by a pre-trained recipe semantics encoder with three main components: key term extraction module, term rating module and term embedding module. 

\paragraph{Key term extraction module} The key terms are broadly divided into ingredients (e.g., pork), cooking utensils (e.g., pan) and actions (e.g., stir). In order to extract the ingredients, we use the bi-directional LSTM networks to analyze the word sequence of each ingredient in the list, such as ``half pound ditali pasta", and learn to identify and extract those word sequences that are ingredients with high probability, like ``ditali pasta". Then we perform binary logistic regression on the extracted word sequence set to produce two clusters, labeled as true or false in the context of core ingredients. Those with true labels are regarded as the ingredient entities to the recipe.
Instead of bluntly removing those word sequences labeled as false in the existing works, we collect all the true ingredient entities in the whole dataset and reexamine every false labeled word sequence if there is any true ingredient entity. If yes, we will also extract this ingredient entity from the false word sequence to make sure that we do not miss any true ingredient entity. Figure~\ref{miss_ingr} shows an illustration of using the true extracted LSTM results to reexamine those word sequences marked as "false", and correct errors of feature learning via LSTM ingredient entity extractor, effectively recovering important ingredient entities that would otherwise be missed, e.g., ``red\_kidney\_beans". As for the cooking utensils and actions, we resort to the part-of-speech tagging approach~\cite{kumawat2015pos} to extract the nouns and verbs in the whole recipe text where the ingredient entities are removed in advance.

\paragraph{Term rating module} After getting the key terms from the key term extraction module, we need to identify the terms that can uniquely distinguish its recipe from other recipes since the discrimination significance contributed by each term to its recipe is different. We achieve this purpose by using a term rating algorithm (e.g., TFIDF~\cite{Salton-1988}, TextRank~\cite{mihalcea2004_textrank} or BERT~\cite{devlin2018_bert} based approach) and assign each key term with a significance weight. TFIDF and TextRank can obtain the representative terms for the recipe considering the occurrence frequency, while BERT based approach mainly focuses on the semantic correlation between each term and its recipe text. Based on the results in Section~\ref{experiments}, using TFIDF approach can get better performance in our work.

\paragraph{Term embedding module} To get the term embedding for each extracted key term, a word2vec model~\cite{Mikolov-2013} is trained over the recipe texts in the training data, based on the Continuous Bags of Words (CBOW) model framework. 
Every word in the training recipe corpus can obtain its 300-dimension word2vec vector through the learned word embedding matrix, which captures the word sequence patterns in the text recipes. By combining the word2vec vectors of all the key terms from the term embedding module, weighted by their significance weight values from the term rating module, we get the recipe key term semantics feature. 

\subsubsection{Image Category Semantics} 
For learning food image embedding, most of the existing methods resort to pre-trained CNN networks as image encoders, e.g., VGG-16~\cite{Simonyan-2014} and ResNet-50~\cite{He+CVPR2016}. Unfortunately, the neural features learned directly from the pixel grid of food image fails to capture the fact that the food images corresponding to a cooking recipe may have different visual appearances due to different ways of using ingredients, different decorations and background. 
We address this problem by exploring semantic knowledge about food images in the recipe training dataset by a pre-trained image semantics encoder to iteratively align the image embedding closer to the associated recipe text embedding in the common latent representation space for learning a high-quality joint embedding. For example, to capture the inherent semantic relationship between recipe text embedding and image embedding, we extract the image category semantics from the food images by leveraging the Food-101 classifier~\cite{Food101} to strengthen the alignment between the recipe text embedding and the food image embedding. 
The pre-trained image semantics encoder consists of two modules: image category classifier module and term embedding module. We pre-train an image category classifier using WideResNet-50~\cite{wideResnet} on the food101 dataset in the {\bf image category semantics module}, which will predict one of the 101 food categories for a food image. The {\bf term embedding module} here is the same as the module in the recipe encoder, which can generate a word2vec representation for the predicted category label from the image category semantics module.

\subsection{Calibrated Joint Embedding Learning}

This Phase II deep feature calibration for joint embedding learning will train a joint embedding model on the entire training recipe set by first leveraging the Phase I calibration to preprocess the raw training data. The joint embedding model learns the recipe text embedding, the image embedding, and the joint embedding loss function concurrently by taking one recipe training data at a time. For each recipe training input, it has the raw text component (title, ingredient list, cooking instructions) and raw image component. The former is preprocessed by the pre-trained models from Phase I calibration to produce (i) the set of key term embeddings, one for each key ingredient, and(ii) the list of LSTM sequence embeddings, one for each cooking instruction sentence, which will be the preprocessed training input for joint embedding learning. Similarly, the raw image component will be preprocessed to produce the image semantic category embedding, which would be fed into the image embedding learning and joint embedding loss function in the Phase II calibration.

\subsubsection{Learning Text Embedding} The text embedding learning takes two types of inputs for each recipe input in the training set: the sequence of sentence embeddings for recipe instruction, and the list of key term embeddings (e.g., in 300 dimensions), weighted by their corresponding key-term rating scores. The former is fed into an LSTM model to learn the recipe instruction embedding, say in the latent representation of 1024 dimensions. The latter is fed into a fully-connected layer to obtain the recipe ingredients embedding in the same latent dimension as the LSTM instruction sequence embedding (e.g., 1024) for the same recipe. Finally, the text embedding for this recipe is learned by taking the concatenation of the embedding for both the key terms and the LSTM-based instruction sentence sequence embedding, and producing the latent recipe text representation in the cross-modal joint embedding space through a fully connected layer.

%%And we assume that the recipe key term semantics feature is very critical for the quality of learned recipe embedding, so we decide to assign the same weight to the key term semantics feature as the recipe sequence feature by feeding the key term semantics feature into a fully-connected layer to obtain the recipe semantics embedding with the same dimension as recipe sequence embedding (1024). Then, recipe semantics embedding and the LSTM based recipe sequence embedding are combined through concatenation and fed into a fully connected layer of the recipe text embedding to produce the 1024-dimension mapping to the same latent space in which the recipe embedding can be compared with the food image embedding.

%%\subsubsection{Image Embedding Learning} 
\subsubsection{Learning Image Embedding}
Similarly, the raw image is prepared in the required pixel resolution and fed into the ResNet50 auto-encoder to learn the image pixel embedding. In addition, the raw image is also preprocessed to produce the image category semantics embedding (e.g., 300 dimensions) using the pre-trained model from Phase I calibration. Both embeddings are finally fed as concatenation to a fully connected layer to output the image embedding for this training recipe, which together with the text embedding for the same recipe will be fed into the learning of joint embedding loss function. 

%%Due to the accuracy limit of image category classifier, we assign a small weight of image category semantics embedding by keeping its dimension (300) small compared to the dimension of ResNet50 auto-encoder generated food image pixel embedding (2048). We integrate this image semantics embedding with the image pixel embedding by vector concatenation, and then followed by a fully-connected layer to get the final image category semantics enhanced image embedding. 

\subsubsection{Joint Embedding Loss Optimization} 

To efficiently optimize the learning of joint embedding loss function, we design the primary loss function by the soft-margin based batch-hard triplet loss empowered with a novel double-negative sampling strategy, denoted as $L_{TRI}$, and further enhanced by incorporating two auxiliary loss regularizations: category-based alignment loss on both cooking recipe and food image $L_{CA}$, and discriminator based alignment loss $L_{DA}$. We define the overall objective function for joint embedding loss optimization as follows: 
\begin{equation} 
L = L_{TRI} + \lambda_1 L_{CA} + \lambda_2  L_{DA} 
\label{global}
\end{equation} 
\noindent  
where $\lambda_1$, $\lambda_2$ are trade-off hyper-parameters.

\paragraph{Double negative sampling and soft-margin optimized batch hard triplet loss} Triplet loss~\cite{Hermans+2017} is calculated on the triplet of training samples $(x_a, x_p, x_n)$, where $x_a$ represents a feature embedding as an anchor point in one modality and is used as the ground truth to evaluate the corresponding modality embedding, $x_p$ and $x_n$ denote the positive and negative feature embeddings from the other modality. The triplet loss optimization is to ensure that the positive instance in one modality should be close to the anchor point in the other modality, and the negative instance in one modality should be distant from the anchor point in the other modality. By selecting the hardest positive and negative samples for each anchor point within every batch when calculating the triplet loss, Hermans~\cite{Hermans+2017} shows that it often outperforms the {\em batch-all triplet loss}, which is based on the average distance from all negative examples to the anchor point. 
In our work, we first optimize the batch-hard triplet loss~\cite{Hermans+2017} by introducing double negative sampling: given an anchor point of one modality and its other modality as the positive sample, among the remaining negative instances in the batch, we select the hardest negative instance using two constraints: closest to the anchor point in the joint embedding space and yet with different category label. 
Second, we use the softplus function $ln(1+\exp(\gamma(\cdot+m)))$ as a smooth approximation to replace the hinge function $[m+\cdot]_+$ used in existing works~\cite{Carvalho+SIGIR2018_AdaMine,JinJinChen+MM2018_AMSR,Hao+CVPR2019_ACME}, which assumes that the distance between the anchor point and negative instance is always larger than the distance between the anchor and positive instance by a fixed margin $m$. Our soft-margin based approach improves the hinge with an exponential decay instead of a threshold-based hard cut-off, which is given as:

\begin{equation}  
\label{TRI}  
\begin{aligned}  
L_{TRI} &= \sum^{N}_{i=1}ln(1+e^{\gamma({d(E_{r_i}^a, E_{v_i}^p)}-\min{d(E_{r_i}^a, E_{v_i}^n)} + m)})  \\ &+ \sum^{N}_{i=1}ln(1+e^{\gamma( d(E_{v_i}^a, E_{r_i}^p)-\min d(E_{v_i}^a, E_{r_i}^n) + m)} )  
 \end{aligned}  
 \end{equation} 

where $d(\cdot)$ measures the Euclidean distance between two input vectors, $N$ is the number of the different recipe-image pairs in a batch, subscripts $a$, $p$ and $n$ refer to anchor, positive and negative instances respectively, $E_{r_i}, E_{v_i}$ refer to the embeddings of the recipe and image in the $i$-th recipe-image pair respectively, $\gamma$ is the scaling factor and $m$ denotes the margin of error in the triplet loss.

 \begin{table*} 
		\center 
		\caption{{Performance comparison of our SEJE with existing representative methods on the 1k and 10k test set. The symbol ``-'' indicates that the results are not available from the corresponding works.}} 
		\label{main_results} 
		\begin{tabular}{cc|cccc|cccc} 
		\hline
		\multirow{2} * {\tabincell{c}{Size of \\test-set}} & \multirow{2} * {Approaches}  &\multicolumn{4}{c}{Image to recipe retrieval} & \multicolumn{4}{c}{Recipe to image retrieval } \\ 
		 \cline{3-10} 
		   ~ & ~ & MedR$\downarrow$ & R@1$\uparrow$ &R@5$\uparrow$ & R@10$\uparrow$ & MedR$\downarrow$ & R@1$\uparrow$ &R@5$\uparrow$ & R@10$\uparrow$ \\   
		\hline
		\multirow{12} *{1k}&  JESR ~\cite{Salvador+CVPR2017_JESR}  & 5.2 & 24.0 & 51.0 & 65.0 & 5.1 & 25.0 & 52.0 & 65.0 \\ 
		~ & Img2img+JESR~\cite{lien2020recipe}  & - & - & - & - & 5.1 & 23.9 & 51.3 & 64.1 \\ 
		~ & AMSR~\cite{JinJinChen+MM2018_AMSR}  & 4.6 & 25.6 & 53.7 & 66.9 & 4.6 & 25.7 & 53.9 & 67.1 \\ 
		~ & AdaMine~\cite{Carvalho+SIGIR2018_AdaMine}  & 1.0 & 39.8 & 69.0 & 77.7 & 1.0 & 40.2 & 68.1 & 78.7 \\ 
		~ & R$^2$GAN~\cite{zhu2019r2gan} &2.0 & 39.1 & 71.0 & 81.7 & 2.0 & 40.6 & 72.6 & 83.3 \\
		~ & MCEN~\cite{fu2020mcen} &2.0 & 48.2 & 75.8 & 83.6 & 1.9 & 48.4 & 76.1 & 83.7 \\
		~ & ACME~\cite{Hao+CVPR2019_ACME} &1.0 & 51.8 & 80.2 & 87.5 & 1.0 & 52.8 & 80.2 & 87.6 \\

		 \cline{2-10} 
		~ & SEJE(TextRank) & 1.0 & 51.9  & 81.5 & 88.9 & 1.0  & 53.0  & 82.1  & 89.1 \\ 
		~ & SEJE(DistilBERT) & 1.0 & 52.3  & 81.6 & 88.6 & 1.0  & 53.6  & 82.0  & 89.2 \\ 
		~ & SEJE(BERT) & 1.0 & 52.8  & 81.7 & 89.2 & 1.0  & 53.7  & 82.3  & 89.5 \\ 
		~ & SEJE(RoBERTa) & 1.0 & 54.6  & 83.3 & 90.4 & 1.0  & 55.2  & 83.7  & 90.7 \\ 
		~ & \textbf{SEJE(TFIDF)} & 1.0 & \textbf{57.2}  & \textbf{85.2} & \textbf{91.2} & 1.0  & \textbf{57.4}  & \textbf{85.7}  & \textbf{91.7} \\ 
		\hline 
		 \multirow{11} *{10k} & JESR~\cite{Salvador+CVPR2017_JESR} & 41.9 & - & - & - & 39.2 & - & - & - \\ 
		~ & AMSR~\cite{JinJinChen+MM2018_AMSR}   & 39.8 & 7.2 & 19.2 & 27.6 & 38.1 & 7.0 & 19.4 & 27.8 \\ 
		~ & AdaMine~\cite{Carvalho+SIGIR2018_AdaMine}  & 13.2 & 14.9 & 35.3 & 45.2 & 12.2 & 14.8 & 34.6 & 46.1 \\ 
		~ & R$^2$GAN~\cite{zhu2019r2gan} &13.9 & 13.5 & 33.5 & 44.9 & 12.6 & 14.2 & 35.0 & 46.8 \\
		~ & MCEN~\cite{fu2020mcen} & 7.2 & 20.3 & 43.3 & 54.4 & 6.6 & 21.4 & 44.3 & 55.2 \\
		~ & ACME~\cite{Hao+CVPR2019_ACME}& 6.7 & 22.9 & 46.8 & 57.9 & 6.0 & 24.4 & 47.9 & 59.0 \\ 
		 \cline{2-10} 
		~ & SEJE(TextRank) & 6.0 & 21.9  & 47.2 & 59.2 & 6.0  & 22.8  & 48.1  & 59.9 \\ 
		~ & SEJE(DistilBERT) & 6.0 & 22.7  & 48.3 & 60.1 & 6.0  & 23.7  & 48.8 & 60.5 \\ 
		~ & SEJE(BERT) & 6.0 & 23.2  & 48.8 & 60.6 & 6.0  & 24.0  & 49.6  & 61.2 \\ 
		~ & SEJE(RoBERTa) & \textbf{5.0} & 24.0  & 50.3 & 62.2 & 5.0  & 24.9  & 50.8  & 62.6 \\
		~ & \textbf{SEJE(TFIDF)} & \textbf{5.0} & \textbf{26.3}  & \textbf{53.2} & \textbf{64.9} & \textbf{4.9}  & \textbf{27.0}  & \textbf{53.7}  & \textbf{65.3}  \\ 
		\hline
		\end{tabular} 
\end{table*} 

\paragraph{Category-based alignment loss}
The category-based loss regularizations on both learned recipe and image embeddings aim to reduce the cross-entropy loss between each of the $N$ modality-specific embeddings and the corresponding category from the total of $N_{c}$ categories obtained from the recipe text and the associated food image (recall $L_{CA-R}$ and $L_{CA-V}$ in Figure~\ref{general_framework}).  
SEJE assigns every recipe-image pair to one of the 1005 category labels learned from text mining analysis on the whole recipe text and the image category classifier module in the Phase I deep feature calibration, 
avoiding assigning background labels to a large percentage of recipe-image pairs, as in existing approaches~\cite{Salvador+CVPR2017_JESR,Carvalho+SIGIR2018_AdaMine,JinJinChen+MM2018_AMSR,Hao+CVPR2019_ACME}. This cross-modal category distribution alignment between the textual recipe and visual image is utilized as a regularization to our primary joint embedding loss optimization $L_{TRI}$. The category-based loss regularization is applied to both image and recipe as follows:

\begin{equation}  
\begin{array}{c}  
L_{CA} = L_{CA-R} + L_{CA-V}\\
L_{CA-R} = -\sum^{N}_{i=1}\sum^{N_c}_{t=1}{y^{i,t}_R \log(\hat{y}^{i,t}_R)}  \\  
L_{CA-V} = -\sum^{N}_{i=1}\sum^{N_c}_{t=1}{y^{i,t}_V \log(\hat{y}^{i,t}_V)}  
 \end{array}  
 \end{equation}
where $L_{CA-R}$ is the loss of regularization on the recipe embedding, while $L_{CA-V}$ is on the image embedding. $N_c$ is the number of category labels, $N$ is the number of the different recipe-image pairs in a batch, $y^{i,t}_R$ and $\hat{y}^{i,t}_R$ are the true and estimated possibilities that the $i^{th}$ recipe embedding belongs to the $t^{th}$ category label, and similarly, $y^{i,t}_{V}$ and $\hat{y}^{i,t}_{V}$ are defined for image embedding.

\begin{figure*}[h]
  \centering  
  \includegraphics[scale=0.23]{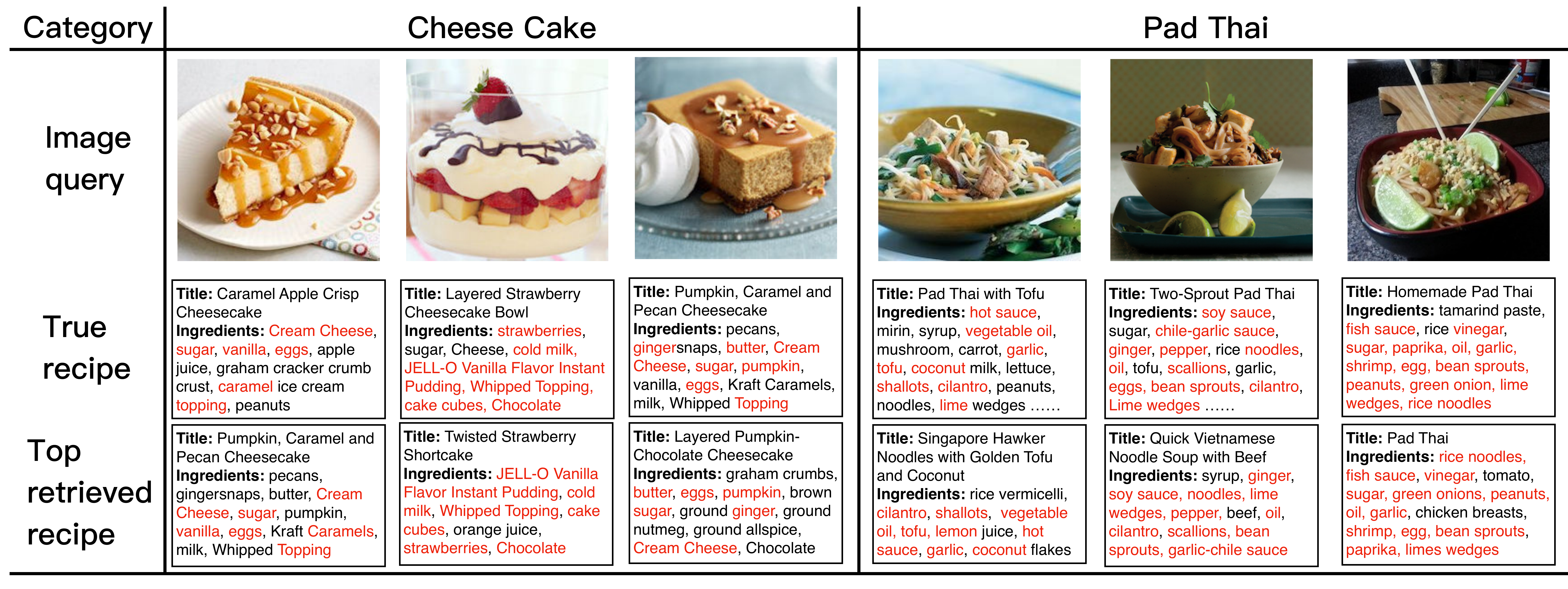}  
  \caption{The results of image-to-recipe retrieval by SEJE approach on the 10k test set. The common or similar ingredients in the true recipe and top retrieved recipe are highlighted in red.}   
  \label{JEMA-i2r results}  
\end{figure*} 

\begin{figure*}[h]
  \centering  
  \includegraphics[scale=0.21]{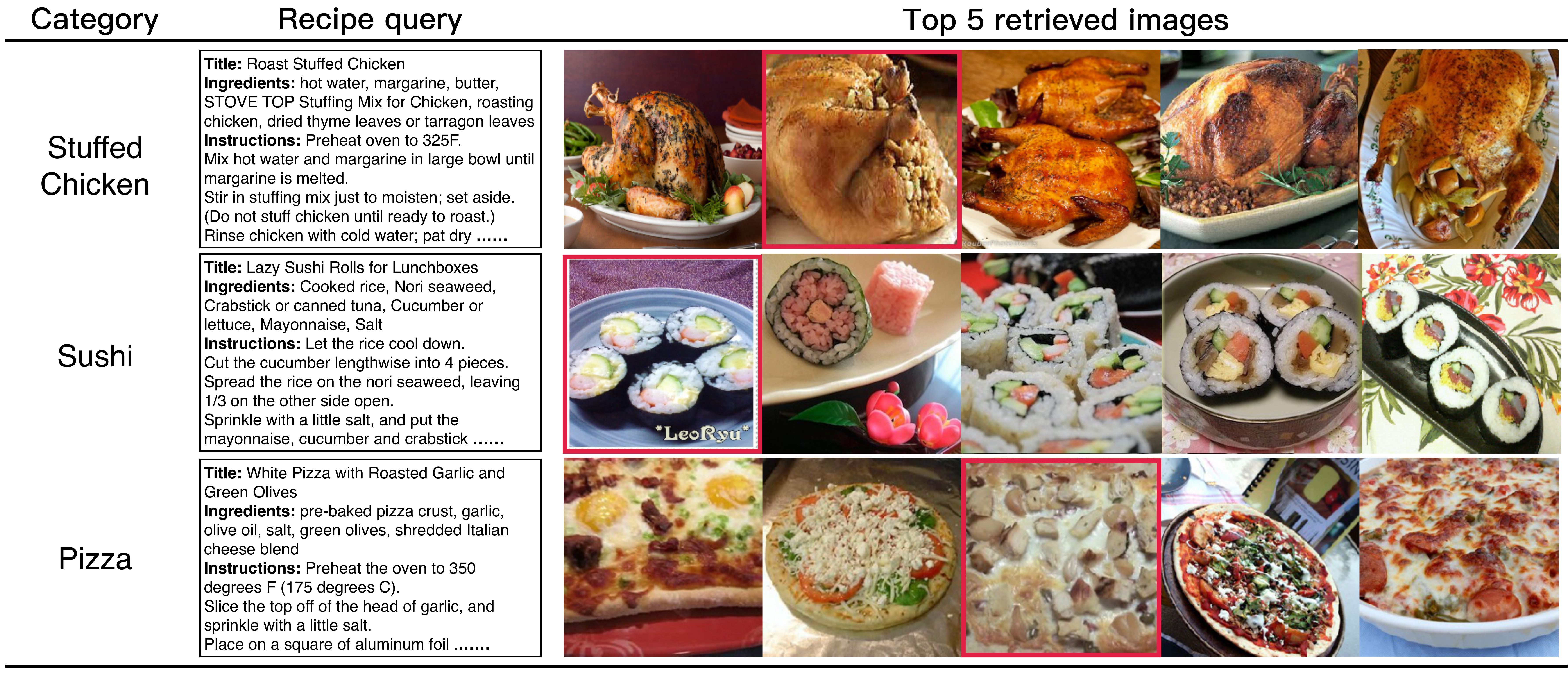}  
  \caption{The results of recipe-to-image retrieval by SEJE approach on the 10k test set. The matched images are boxed in red.}   
  \label{JEMA-r2i results}  
\end{figure*} 

\paragraph{Discriminator based alignment loss}
To further reduce possible errors inherent in our primary triplet loss ($L_{TRI}$), we also use the discriminator based alignment loss as another regularization, which utilizes the competing strategy like those in GAN~\cite{goodfellow+2014} with the gradient penalty~\cite{gulrajani+2017}. Concretely, as a part of the joint embedding learning, we train a discriminator model such that given an embedding from one modality, this discriminator can tell whether it is from the image or the recipe text, which is against the goal of a good-quality joint embedding learning. If the trained discriminator cannot accurately discriminate the embedding of one modality from the embedding of the other modality, it indicates that our joint embedding learning is effective and the learned embedding distributions for matched recipe-image pairs are too close in their joint latent space $\mathbb{R}^d$ for the discriminator to tell them apart. Since the recipe and image embeddings play the same role during the discriminator learning, we select the image embedding as the target of the learned discriminator, which means receiving image embeddings, the learned discriminator would give a high confidence value while recipe embedding would result in low confidence value. The higher confidence values for recipe embeddings are obtained by our image discriminator (note that the confidence values of image embeddings do not need to be low), the more aligned the distributions of recipe embeddings and image embeddings are in the latent embedding space $\mathbb{R}^d$. The discriminator networks used here are made up of three fully-connected layers. The loss for the discriminator networks $L_D$ and the cross-modal discriminator based alignment loss are given as:

\begin{equation} 
\begin{aligned} 
L_D&=\sum^{N}_{i=1}[\log(F_D(E_R(r_i)))+\log(1-F_D(E_V(v_i)))\\
 &+\lambda_D(\Vert \nabla_{x_i}F_D(x_i)\Vert_2-1)^2] 
 \end{aligned} 
 \label{D}
 \end{equation} 
\begin{equation} 
L_{DA} = \sum^{N}_{i=1}{\log(1-F_D(E_R(r_i)))}  
\end{equation} 

\noindent 
where $F_D(\cdot)$ is the function of our trained discriminator, whose output is the confidence value that it distinguishes the input embedding as the image embedding, $N$ is the number of the different recipe-image pairs in a batch, $\lambda_D$ is the trade-off parameter, $x_i$ is a random interpolation between the $i^{th}$ recipe embedding $E_R(r_i)$ and image embedding $E_V(v_i)$.

\begin{table}[t]
		\center
		\caption{Evaluation of contributions of different components in the SEJE framework for the image-to-recipe retrieval on the 1k test-set.
}
		\label{ablation results}
		\begin{tabular}{c|cccc}
		\hline
		 \multirow{2} * {Component} & \multicolumn{4}{c}{Image to recipe retrieval } \\
		 \cline{2-5}
		   ~ & MedR & R@1 &R@5 & R@10  \\  
		\hline
		SEJE-b & 4.1 & 25.9 & 56.4 & 70.1 \\
		\hline
		SEJE-b+Cal$_{V}$& 3.4 & 28.1 & 59.5 & 73.1  \\
		SEJE-b+Cal$_R$& 3.0 & 29.4 & 60.0 & 73.4 \\		
		SEJE-b+P I & 3.0 & 30.5 & 61.6 & 75.2 \\
		\hline
		SEJE-b+P I+CA(-) & 2.5 & 32.6 & 68.2 & 80.3\\
		SEJE-b+P I+CA & 2.3 & 33.5 & 68.4 & 80.9 \\
		SEJE-b+P I+DA & 2.5 & 36.0 & 65.2 & 77.3 \\
		SEJE-b+P I+TRI(-) & 2.0 & 46.5 & 78.1 & 86.7\\
		SEJE-b+P I+TRI & 1.6 & 47.7 & 78.6 & 87.3 \\ 
		SEJE-b+P I+ P II & \textbf{1.0}  & \textbf{57.2}  & \textbf{85.2} & \textbf{91.2}\\  	
		\hline
		\end{tabular}
\end{table}

\begin{table*} [h]
		\center 
		\caption{{Evaluation of contributions of each component in the extracted key terms on the 10k test-set.}} 
		\label{ablation study} 
		\begin{tabular}{c|cccc|cccc} 
		\hline
		 \multirow{2} * {Components}  &\multicolumn{4}{c}{Image to recipe retrieval} & \multicolumn{4}{c}{Recipe to image retrieval } \\ 
		 \cline{2-9} 
		   ~ & MedR$\downarrow$ & R@1$\uparrow$ &R@5$\uparrow$ & R@10$\uparrow$ & MedR$\downarrow$ & R@1$\uparrow$ &R@5$\uparrow$ & R@10$\uparrow$ \\   
		\hline
		 ingredient & 5.0 & 25.0  & 51.4 & 63.1 & 5.0  & 26.1  & 52.0  & 63.4 \\
		 ingredient+utensil & 5.0 & 25.6  & 52.2 & 64.1 & 5.0  & 26.4  & 52.5  & 64.2 \\ 
		 ingredient+utensil+action & 5.0  & \textbf{26.3}  & \textbf{53.2} & \textbf{64.9} & \textbf{4.9}  & \textbf{27.0}  & \textbf{53.7}  & \textbf{65.3} \\ 
		\hline 
		\end{tabular} 
\end{table*}

\section{Experiments}
\label{experiments}

\paragraph{Dataset and Evaluation Metrics.}
We evaluate the effectiveness of different approaches on the Recipe1M dataset, consisting of over 800K recipes (title, list of ingredients and cooking instructions) and 1 million associated food images.
Experiment setup follows the literature: (i) Sample 10 unique subsets of 1k or 10k matching recipe-image pairs from the test set. (ii) Use each item in one modality as a query (e.g., an image), and rank instances in the other modality (e.g., recipes) by the Euclidean distance between the query embedding and each candidate embedding from the other modality in the test set. (iii) Calculate the median retrieval rank (MedR) and the recall percentage at top K (R@K), i.e., the percentage of queries for which the matching answer is included in the top $K$ results ($K$=1,5,10). Note that even though many loss regularizations are added in our framework, these optimizations are only performed at the training stage but not in the testing stage. Seven baselines are considered: JESR, Img2img+JESR, AMSR, AdaMine, R$^2$GAN, ACME and MCEN. 

\paragraph{ Implementation Details.} 
Public word2vec toolkit~\footnote{https://code.google.com/archive/p/word2vec/} is used to train the Continuous Bag-of-Words (CBOW) version of word2vec model~\cite{word2vec-2013} on the corpus of recipe texts in the Recipe1M dataset. The part-of-speech tagging used in this work is implemented by using the public Python library TextBlob~\footnote{https://textblob.readthedocs.io/en/dev/}. The dimensions of joint embedding and word2vec embedding are set as 1024 and 300 respectively. Since these hyper-parameters are very critical for the final performance, we have experimentally evaluated their impacts and made an observation that the best results can be obtained when $\lambda_1$, $\lambda_2$ in Equation~\ref{global}, $\gamma$, $m$ in Equation~\ref{TRI}, $\lambda_D$ in Equation~\ref{D} are set as 0.005, 0.005, 16, 0.0 and 10 respectively. Adam optimizer~\cite{Adam-2014} is employed for model training with the initial learning rate set as $10^{-4}$ in all experiments, with the mini bath size of 100. All deep neural networks are implemented on the Pytorch platform and trained on a single Nvidia Titan X Pascal server with 12GB of memory. When using the trained cross-modal embedding model to perform recipe-image cross-modal retrieval, it takes only about 0.13 seconds to perform a cross-modal retrieval on the 1k test set, and about 1.5 seconds on the 10k test set.

\subsection{Cross-Modal Retrieval Performance}
\noindent
We evaluate the performance of our SEJE approach for image-to-recipe and recipe-to-image retrieval against baselines in Table~\ref{main_results}. The results of JESR, Img2img+JESR, AMSR, AdaMine, R$^2$GAN, ACME and MCEN are from ~\cite{Salvador+CVPR2017_JESR,lien2020recipe,JinJinChen+MM2018_AMSR,Carvalho+SIGIR2018_AdaMine,zhu2019r2gan,Hao+CVPR2019_ACME,fu2020mcen} for the fair comparison. We observe that SEJE consistently outperforms all baselines for both image-to-recipe and recipe-to-image queries on 1k and 10k test data, showing the effectiveness of using the deep feature calibration techniques to incorporate the additional semantics for further optimizing the cross-modal joint embedding learning. We also evaluate different term rating algorithms used in the term rating module: TFIDF, TextRank and three popular BERT models, which are standard BERT~\cite{devlin2018_bert}, DistilBERT~\cite{sanh2019_distilbert} and RoBERTa~\cite{liu2019_roberta}. The results are reported in Table~\ref{main_results} under SEJE sections, which indicates that selecting the TFIDF approach in the term rating module can yield the best accuracy in our work.

\begin{table*} [h]
		\center 
		\caption{{Performance comparison of our SEJE using different key term filters on the 10k test set. The symbol ``-'' indicates that the results don't use the key term filter.}} 
		\label{weight threshold} 
		\begin{tabular}{cc|cccc|cccc} 
		\hline
		\multirow{2} * {\tabincell{c}{Term Rating \\Algorithm}} & \multirow{2} * {Threshold}  &\multicolumn{4}{c}{Image to recipe retrieval} & \multicolumn{4}{c}{Recipe to image retrieval } \\ 
		 \cline{3-10} 
		   ~ & ~ & MedR$\downarrow$ & R@1$\uparrow$ &R@5$\uparrow$ & R@10$\uparrow$ & MedR$\downarrow$ & R@1$\uparrow$ &R@5$\uparrow$ & R@10$\uparrow$ \\   
		\hline
		\multirow{4} *{RoBERT} & - & 5.0 & 24.0  & 50.3 & 62.2 & 5.0  & 24.9  & 50.8  & 62.6 \\
		~ & 0.05 & 5.0 & 25.1  & 51.7 & 63.5 & 5.0  & 25.8  & 52.3  & 64.0 \\ 
		~ & 0.10 & 5.0 & 24.2  & 50.4 & 62.5 & 5.0  & 25.3  & 51.2  & 62.7 \\ 
		~ & 0.15 & 5.0 & 24.0  & 50.3 & 62.0 & 5.0  & 25.0  & 50.8  & 62.6 \\ 
		\hline 
		 \multirow{4} *{TFIDF} & - & 5.0 & \textbf{26.3}  & \textbf{53.2} & \textbf{64.9} & \textbf{4.9}  & \textbf{27.0}  & \textbf{53.7}  & \textbf{65.3}  \\ 
		~ & 0.05 & 5.0 & 26.1  & 53.1 & 64.8 & 5.0  & 26.7  & 53.7 & \textbf{65.2} \\ 
		~ & 0.10 & 5.0 & 25.3  & 51.7 & 63.4 & 5.0  & 26.1  & 52.5  & 64.1 \\ 
		~ & 0.15 & 6.0 & 22.7  & 48.0 & 59.9 & 6.0  & 23.5  & 49.1  & 60.4 \\
		\hline
		\end{tabular} 
\end{table*} 

\subsection{Visualization Results}
\noindent We provide some visualization results of SEJE for both image-to-recipe and recipe-to-image retrieval tasks on the 10k dataset in Figure~\ref{JEMA-i2r results} and Figure~\ref{JEMA-r2i results} respectively. 
Figure~\ref{JEMA-i2r results} shows the ground truth recipes and the top retrieved recipes based on 6 different image queries. We consider two recipe categories: \textit{cheese cake} and \textit{pad thai}, and in each category we list three image queries. It shows that almost all the significant ingredients of the true recipes and visual components in the image queries also appear in the top retrieved recipes. Figure~\ref{JEMA-r2i results} visualizes the results of retrieving the top 5 images using three different recipe queries. In all cases, most of the retrieved images share similar ingredients to the ground truth images and contain the visual components as suggested by the category label.
In the first example, all the retrieved images contain the visual component of stuffed chicken, as suggested by the category label. In the second example, all the top 5 results can be recognized as sushi and visually similar to the ground truth image. All the retrieved images in the third example are visually similar and all are about pizza. 
Figure~\ref{JEMA-i2r results} and Figure~\ref{JEMA-r2i results} further illustrate the effectiveness of our SEJE approach for text-image joint embedding learning and the two-phase deep feature calibration can boost the accuracy of the cross-modal retrieval task.

\subsection{Ablation Study}

\noindent
We evaluate the contributions of each core component in our two-phase deep feature calibrated SEJE approach using image-to-recipe retrieval. Table~\ref{ablation results} reports the results. We use \textbf{SEJE-b} to denote the naive SEJE baseline with the simple batch-all triplet loss and without our proposed phase I preprocessing. We then incrementally add one component at a time. First, we analyze the gains from employing the semantic calibration on image and recipe to leverage the additional image and recipe semantics respectively, i.e., \textbf{Cal$_V$} and \textbf{Cal$_R$}, and the text calibration incorporates the term rating algorithm by TFIDF. We use \textbf{P I} to represent that the phase I deep feature calibration techniques for both image and text are employed. Then we add the category-based alignment loss regularizations \textbf{CA} on both recipe and image embeddings, and we also list the results of employing the category-based alignment loss with the original category labels proposed in JESR (denoted as \textbf{CA(-)}). Next, we add the cross-modal discriminator-based alignment loss regularization \textbf{DA}. Then we replace the simple batch-all triplet loss in \textbf{SEJE-b} with our double negative sampling and soft-margin optimized batch-hard triplet loss, denoted by \textbf{TRI}. And we also report the results of employing the standard batch-hard triplet loss (denoted as \textbf{TRI(-)}) to compare with our optimized batch-hard triplet loss. Finally, we add the complete phase II deep feature calibrated joint embedding, coined \textbf{P II}. Table~\ref{ablation results} shows that each proposed component positively contributes towards improving the cross-modal alignment between recipe text and food image, effectively boosting the overall performance of cross-modal retrieval.

\subsection{ Effect of Term Extraction and Ranking}
As claimed in the introduction of the key term extraction module, the extracted key terms from the recipe text during the data preprocessing can be broadly divided into three types: ingredient entities, cooking utensils and actions. In this section, we want to investigate the significance of employing term extraction and term rating score on each type of key terms on the overall performance of SEJE.

\subsubsection{Evaluation of Each Key Term Component}
Here we want to evaluate the contributions of each core component in our extracted key terms. Specifically, we set our SEJE approach that only extracts the ingredients and adopts the TFIDF approach as the term rating algorithm as the bare bone baseline. Then we incrementally add one component of the cooking utensils and actions at a time. Table~\ref{ablation study} reports the results. It shows that every proposed component positively contributes towards improving the cross-modal alignment between recipe text and food image and boosting the overall performance of cross-modal retrieval. 

\subsubsection{Evaluation of the Key Term Filter}
In order to capture the level of discrimination significance contributed by each key term to its recipe text, we assign different weight to each key term extracted from the recipe text by several term rating algorithms (i.e. TextRank, BERT based approach and TFIDF). The more discriminative the key term is for its recipe, the higher weight it should be given. Since the terms with very low weights might be useless and even harmful to the quality of the learned recipe embedding, we conduct several experiments where the terms with weights lower than a threshold $k$ are removed when generating the recipe key term semantics feature. Recall the experimental results of using different term rating algorithms in Table~\ref{main_results}, we can find that using the RoBERT model and TFIDF approach can get better performance. Therefore, we validate our assumption that using the key term filter to remove those key terms with low weight might improve the quality of the recipe embedding learning based on the RoBERT model and TFIDF approach. The results are reported in Table~\ref{weight threshold}, which indicate that using the key term filter on the RoBERT based term rating algorithm can boost the performance when the weight threshold is set as 0.05, while the key term filter can not benefit the TFIDF based term rating algorithm.

\section{Conclusion}

We have presented SEJE, a two-phase deep feature calibration framework for learning cross-modality joint embedding with the semantics enhanced feature embeddings and loss optimizations. Our SEJE method can extract and incorporate the additional semantics in both cooking recipe text and food image to capture more discriminative properties of input recipe text and its associated food image, and semantically align the learned recipe and image embeddings. By further integrated with soft-margin optimized batch-hard triplet loss with the double negative sampling strategy as the primary loss and the category-based alignment loss and discriminative-based alignment loss as the two auxiliary loss regularizations, SEJE effectively boosts the accuracy and retrieval performance of cross-modal joint embedding learning and it outperforms the seven representative state-of-the-art methods for both image-to-recipe and recipe-to-image retrieval on Recipe1M benchmark dataset.   

\begin{acks}
This work is partially supported by the USA National Science Foundation under Grants NSF 2038029, 1564097, and an IBM faculty award. The first author has performed this work as a two-year visiting PhD student at Georgia Institute of Technology (2019-2021), under the support from China Scholarship Council (CSC) and Wuhan University of Technology.
\end{acks}

%%
%% The next two lines define the bibliography style to be used, and
%% the bibliography file.
\bibliographystyle{ACM-Reference-Format}
\bibliography{ICMI-camera-ready}

%%% -*-BibTeX-*-
%%% Do NOT edit. File created by BibTeX with style
%%% ACM-Reference-Format-Journals [18-Jan-2012].

\begin{thebibliography}{26}

%%% ====================================================================
%%% NOTE TO THE USER: you can override these defaults by providing
%%% customized versions of any of these macros before the \bibliography
%%% command.  Each of them MUST provide its own final punctuation,
%%% except for \shownote{}, \showDOI{}, and \showURL{}.  The latter two
%%% do not use final punctuation, in order to avoid confusing it with
%%% the Web address.
%%%
%%% To suppress output of a particular field, define its macro to expand
%%% to an empty string, or better, \unskip, like this:
%%%
%%% \newcommand{\showDOI}[1]{\unskip}   % LaTeX syntax
%%%
%%% \def \showDOI #1{\unskip}           % plain TeX syntax
%%%
%%% ====================================================================

\ifx \showCODEN    \undefined \def \showCODEN     #1{\unskip}     \fi
\ifx \showDOI      \undefined \def \showDOI       #1{#1}\fi
\ifx \showISBNx    \undefined \def \showISBNx     #1{\unskip}     \fi
\ifx \showISBNxiii \undefined \def \showISBNxiii  #1{\unskip}     \fi
\ifx \showISSN     \undefined \def \showISSN      #1{\unskip}     \fi
\ifx \showLCCN     \undefined \def \showLCCN      #1{\unskip}     \fi
\ifx \shownote     \undefined \def \shownote      #1{#1}          \fi
\ifx \showarticletitle \undefined \def \showarticletitle #1{#1}   \fi
\ifx \showURL      \undefined \def \showURL       {\relax}        \fi
% The following commands are used for tagged output and should be
% invisible to TeX
\providecommand\bibfield[2]{#2}
\providecommand\bibinfo[2]{#2}
\providecommand\natexlab[1]{#1}
\providecommand\showeprint[2][]{arXiv:#2}

\bibitem[\protect\citeauthoryear{Bossard, Guillaumin, and Van~Gool}{Bossard
  et~al\mbox{.}}{2014}]%
        {Food101}
\bibfield{author}{\bibinfo{person}{Lukas Bossard}, \bibinfo{person}{Matthieu
  Guillaumin}, {and} \bibinfo{person}{Luc Van~Gool}.}
  \bibinfo{year}{2014}\natexlab{}.
\newblock \showarticletitle{Food-101--mining discriminative components with
  random forests}. In \bibinfo{booktitle}{\emph{European conference on computer
  vision}}. Springer, \bibinfo{pages}{446--461}.
\newblock


\bibitem[\protect\citeauthoryear{Carvalho, Cad{\`e}ne, Picard, Soulier, Thome,
  and Cord}{Carvalho et~al\mbox{.}}{2018}]%
        {Carvalho+SIGIR2018_AdaMine}
\bibfield{author}{\bibinfo{person}{Micael Carvalho}, \bibinfo{person}{R{\'e}mi
  Cad{\`e}ne}, \bibinfo{person}{David Picard}, \bibinfo{person}{Laure Soulier},
  \bibinfo{person}{Nicolas Thome}, {and} \bibinfo{person}{Matthieu Cord}.}
  \bibinfo{year}{2018}\natexlab{}.
\newblock \showarticletitle{Cross-modal retrieval in the cooking context:
  Learning semantic text-image embeddings}. In \bibinfo{booktitle}{\emph{The
  41st International ACM SIGIR Conference on Research \& Development in
  Information Retrieval}}. \bibinfo{pages}{35--44}.
\newblock


\bibitem[\protect\citeauthoryear{Chen, Ngo, Feng, and Chua}{Chen
  et~al\mbox{.}}{2018}]%
        {JinJinChen+MM2018_AMSR}
\bibfield{author}{\bibinfo{person}{Jing-Jing Chen}, \bibinfo{person}{Chong-Wah
  Ngo}, \bibinfo{person}{Fu-Li Feng}, {and} \bibinfo{person}{Tat-Seng Chua}.}
  \bibinfo{year}{2018}\natexlab{}.
\newblock \showarticletitle{Deep understanding of cooking procedure for
  cross-modal recipe retrieval}. In \bibinfo{booktitle}{\emph{Proceedings of
  the 26th ACM international conference on Multimedia}}.
  \bibinfo{pages}{1020--1028}.
\newblock


\bibitem[\protect\citeauthoryear{Devlin, Chang, Lee, and Toutanova}{Devlin
  et~al\mbox{.}}{2018}]%
        {devlin2018_bert}
\bibfield{author}{\bibinfo{person}{Jacob Devlin}, \bibinfo{person}{Ming-Wei
  Chang}, \bibinfo{person}{Kenton Lee}, {and} \bibinfo{person}{Kristina
  Toutanova}.} \bibinfo{year}{2018}\natexlab{}.
\newblock \showarticletitle{Bert: Pre-training of deep bidirectional
  transformers for language understanding}.
\newblock \bibinfo{journal}{\emph{arXiv preprint arXiv:1810.04805}}
  (\bibinfo{year}{2018}).
\newblock


\bibitem[\protect\citeauthoryear{Fu, Wu, Liu, and Sun}{Fu
  et~al\mbox{.}}{2020}]%
        {fu2020mcen}
\bibfield{author}{\bibinfo{person}{Han Fu}, \bibinfo{person}{Rui Wu},
  \bibinfo{person}{Chenghao Liu}, {and} \bibinfo{person}{Jianling Sun}.}
  \bibinfo{year}{2020}\natexlab{}.
\newblock \showarticletitle{MCEN: Bridging Cross-Modal Gap between Cooking
  Recipes and Dish Images with Latent Variable Model}. In
  \bibinfo{booktitle}{\emph{Proceedings of the IEEE/CVF Conference on Computer
  Vision and Pattern Recognition}}. \bibinfo{pages}{14570--14580}.
\newblock


\bibitem[\protect\citeauthoryear{Goodfellow, Pouget-Abadie, Mirza, Xu,
  Warde-Farley, Ozair, Courville, and Bengio}{Goodfellow et~al\mbox{.}}{2014}]%
        {goodfellow+2014}
\bibfield{author}{\bibinfo{person}{Ian Goodfellow}, \bibinfo{person}{Jean
  Pouget-Abadie}, \bibinfo{person}{Mehdi Mirza}, \bibinfo{person}{Bing Xu},
  \bibinfo{person}{David Warde-Farley}, \bibinfo{person}{Sherjil Ozair},
  \bibinfo{person}{Aaron Courville}, {and} \bibinfo{person}{Yoshua Bengio}.}
  \bibinfo{year}{2014}\natexlab{}.
\newblock \showarticletitle{Generative adversarial nets}. In
  \bibinfo{booktitle}{\emph{Advances in neural information processing
  systems}}. \bibinfo{pages}{2672--2680}.
\newblock


\bibitem[\protect\citeauthoryear{Gulrajani, Ahmed, Arjovsky, Dumoulin, and
  Courville}{Gulrajani et~al\mbox{.}}{2017}]%
        {gulrajani+2017}
\bibfield{author}{\bibinfo{person}{Ishaan Gulrajani}, \bibinfo{person}{Faruk
  Ahmed}, \bibinfo{person}{Martin Arjovsky}, \bibinfo{person}{Vincent
  Dumoulin}, {and} \bibinfo{person}{Aaron~C Courville}.}
  \bibinfo{year}{2017}\natexlab{}.
\newblock \showarticletitle{Improved training of wasserstein gans}. In
  \bibinfo{booktitle}{\emph{Advances in neural information processing
  systems}}. \bibinfo{pages}{5767--5777}.
\newblock


\bibitem[\protect\citeauthoryear{He, Zhang, Ren, and Sun}{He
  et~al\mbox{.}}{2016}]%
        {He+CVPR2016}
\bibfield{author}{\bibinfo{person}{Kaiming He}, \bibinfo{person}{Xiangyu
  Zhang}, \bibinfo{person}{Shaoqing Ren}, {and} \bibinfo{person}{Jian Sun}.}
  \bibinfo{year}{2016}\natexlab{}.
\newblock \showarticletitle{Deep residual learning for image recognition}. In
  \bibinfo{booktitle}{\emph{Proceedings of the IEEE conference on computer
  vision and pattern recognition}}. \bibinfo{pages}{770--778}.
\newblock


\bibitem[\protect\citeauthoryear{Hermans, Beyer, and Leibe}{Hermans
  et~al\mbox{.}}{2017}]%
        {Hermans+2017}
\bibfield{author}{\bibinfo{person}{Alexander Hermans}, \bibinfo{person}{Lucas
  Beyer}, {and} \bibinfo{person}{Bastian Leibe}.}
  \bibinfo{year}{2017}\natexlab{}.
\newblock \showarticletitle{In defense of the triplet loss for person
  re-identification}.
\newblock \bibinfo{journal}{\emph{arXiv preprint arXiv:1703.07737}}
  (\bibinfo{year}{2017}).
\newblock


\bibitem[\protect\citeauthoryear{Kingma and Ba}{Kingma and Ba}{2014}]%
        {Adam-2014}
\bibfield{author}{\bibinfo{person}{Diederik~P Kingma} {and}
  \bibinfo{person}{Jimmy Ba}.} \bibinfo{year}{2014}\natexlab{}.
\newblock \showarticletitle{Adam: A method for stochastic optimization}.
\newblock \bibinfo{journal}{\emph{arXiv preprint arXiv:1412.6980}}
  (\bibinfo{year}{2014}).
\newblock


\bibitem[\protect\citeauthoryear{Kiros, Zhu, Salakhutdinov, Zemel, Urtasun,
  Torralba, and Fidler}{Kiros et~al\mbox{.}}{2015}]%
        {Kiro-2015}
\bibfield{author}{\bibinfo{person}{Ryan Kiros}, \bibinfo{person}{Yukun Zhu},
  \bibinfo{person}{Russ~R Salakhutdinov}, \bibinfo{person}{Richard Zemel},
  \bibinfo{person}{Raquel Urtasun}, \bibinfo{person}{Antonio Torralba}, {and}
  \bibinfo{person}{Sanja Fidler}.} \bibinfo{year}{2015}\natexlab{}.
\newblock \showarticletitle{Skip-thought vectors}. In
  \bibinfo{booktitle}{\emph{Advances in neural information processing
  systems}}. \bibinfo{pages}{3294--3302}.
\newblock


\bibitem[\protect\citeauthoryear{Kumawat and Jain}{Kumawat and Jain}{2015}]%
        {kumawat2015pos}
\bibfield{author}{\bibinfo{person}{Deepika Kumawat} {and}
  \bibinfo{person}{Vinesh Jain}.} \bibinfo{year}{2015}\natexlab{}.
\newblock \showarticletitle{POS tagging approaches: A comparison}.
\newblock \bibinfo{journal}{\emph{International Journal of Computer
  Applications}} \bibinfo{volume}{118}, \bibinfo{number}{6}
  (\bibinfo{year}{2015}).
\newblock


\bibitem[\protect\citeauthoryear{Lien, Zamani, and Croft}{Lien
  et~al\mbox{.}}{2020}]%
        {lien2020recipe}
\bibfield{author}{\bibinfo{person}{Yen-Chieh Lien}, \bibinfo{person}{Hamed
  Zamani}, {and} \bibinfo{person}{W~Bruce Croft}.}
  \bibinfo{year}{2020}\natexlab{}.
\newblock \showarticletitle{Recipe Retrieval with Visual Query of Ingredients}.
  In \bibinfo{booktitle}{\emph{Proceedings of the 43rd International ACM SIGIR
  Conference on Research and Development in Information Retrieval}}.
  \bibinfo{pages}{1565--1568}.
\newblock


\bibitem[\protect\citeauthoryear{Liu, Ott, Goyal, Du, Joshi, Chen, Levy, Lewis,
  Zettlemoyer, and Stoyanov}{Liu et~al\mbox{.}}{2019}]%
        {liu2019_roberta}
\bibfield{author}{\bibinfo{person}{Yinhan Liu}, \bibinfo{person}{Myle Ott},
  \bibinfo{person}{Naman Goyal}, \bibinfo{person}{Jingfei Du},
  \bibinfo{person}{Mandar Joshi}, \bibinfo{person}{Danqi Chen},
  \bibinfo{person}{Omer Levy}, \bibinfo{person}{Mike Lewis},
  \bibinfo{person}{Luke Zettlemoyer}, {and} \bibinfo{person}{Veselin
  Stoyanov}.} \bibinfo{year}{2019}\natexlab{}.
\newblock \showarticletitle{Roberta: A robustly optimized bert pretraining
  approach}.
\newblock \bibinfo{journal}{\emph{arXiv preprint arXiv:1907.11692}}
  (\bibinfo{year}{2019}).
\newblock


\bibitem[\protect\citeauthoryear{Mihalcea and Tarau}{Mihalcea and
  Tarau}{2004}]%
        {mihalcea2004_textrank}
\bibfield{author}{\bibinfo{person}{Rada Mihalcea} {and} \bibinfo{person}{Paul
  Tarau}.} \bibinfo{year}{2004}\natexlab{}.
\newblock \showarticletitle{Textrank: Bringing order into text}. In
  \bibinfo{booktitle}{\emph{Proceedings of the 2004 conference on empirical
  methods in natural language processing}}. \bibinfo{pages}{404--411}.
\newblock


\bibitem[\protect\citeauthoryear{Mikolov, Chen, Corrado, and Dean}{Mikolov
  et~al\mbox{.}}{2013a}]%
        {Mikolov-2013}
\bibfield{author}{\bibinfo{person}{Tomas Mikolov}, \bibinfo{person}{Kai Chen},
  \bibinfo{person}{Greg Corrado}, {and} \bibinfo{person}{Jeffrey Dean}.}
  \bibinfo{year}{2013}\natexlab{a}.
\newblock \showarticletitle{Efficient estimation of word representations in
  vector space}.
\newblock \bibinfo{journal}{\emph{arXiv preprint arXiv:1301.3781}}
  (\bibinfo{year}{2013}).
\newblock


\bibitem[\protect\citeauthoryear{Mikolov, Sutskever, Chen, Corrado, and
  Dean}{Mikolov et~al\mbox{.}}{2013b}]%
        {word2vec-2013}
\bibfield{author}{\bibinfo{person}{Tomas Mikolov}, \bibinfo{person}{Ilya
  Sutskever}, \bibinfo{person}{Kai Chen}, \bibinfo{person}{Greg~S Corrado},
  {and} \bibinfo{person}{Jeff Dean}.} \bibinfo{year}{2013}\natexlab{b}.
\newblock \showarticletitle{Distributed representations of words and phrases
  and their compositionality}. In \bibinfo{booktitle}{\emph{Advances in neural
  information processing systems}}. \bibinfo{pages}{3111--3119}.
\newblock


\bibitem[\protect\citeauthoryear{Salton and Buckley}{Salton and
  Buckley}{1988}]%
        {Salton-1988}
\bibfield{author}{\bibinfo{person}{Gerard Salton} {and}
  \bibinfo{person}{Christopher Buckley}.} \bibinfo{year}{1988}\natexlab{}.
\newblock \showarticletitle{Term-weighting approaches in automatic text
  retrieval}.
\newblock \bibinfo{journal}{\emph{Information processing \& management}}
  \bibinfo{volume}{24}, \bibinfo{number}{5} (\bibinfo{year}{1988}),
  \bibinfo{pages}{513--523}.
\newblock


\bibitem[\protect\citeauthoryear{Salvador, Hynes, Aytar, Marin, Ofli, Weber,
  and Torralba}{Salvador et~al\mbox{.}}{2017}]%
        {Salvador+CVPR2017_JESR}
\bibfield{author}{\bibinfo{person}{Amaia Salvador}, \bibinfo{person}{Nicholas
  Hynes}, \bibinfo{person}{Yusuf Aytar}, \bibinfo{person}{Javier Marin},
  \bibinfo{person}{Ferda Ofli}, \bibinfo{person}{Ingmar Weber}, {and}
  \bibinfo{person}{Antonio Torralba}.} \bibinfo{year}{2017}\natexlab{}.
\newblock \showarticletitle{Learning cross-modal embeddings for cooking recipes
  and food images}. In \bibinfo{booktitle}{\emph{Proceedings of the IEEE
  conference on computer vision and pattern recognition}}.
  \bibinfo{pages}{3020--3028}.
\newblock


\bibitem[\protect\citeauthoryear{Sanh, Debut, Chaumond, and Wolf}{Sanh
  et~al\mbox{.}}{2019}]%
        {sanh2019_distilbert}
\bibfield{author}{\bibinfo{person}{Victor Sanh}, \bibinfo{person}{Lysandre
  Debut}, \bibinfo{person}{Julien Chaumond}, {and} \bibinfo{person}{Thomas
  Wolf}.} \bibinfo{year}{2019}\natexlab{}.
\newblock \showarticletitle{DistilBERT, a distilled version of BERT: smaller,
  faster, cheaper and lighter}.
\newblock \bibinfo{journal}{\emph{arXiv preprint arXiv:1910.01108}}
  (\bibinfo{year}{2019}).
\newblock


\bibitem[\protect\citeauthoryear{Simonyan and Zisserman}{Simonyan and
  Zisserman}{2014}]%
        {Simonyan-2014}
\bibfield{author}{\bibinfo{person}{Karen Simonyan} {and}
  \bibinfo{person}{Andrew Zisserman}.} \bibinfo{year}{2014}\natexlab{}.
\newblock \showarticletitle{Very deep convolutional networks for large-scale
  image recognition}.
\newblock \bibinfo{journal}{\emph{arXiv preprint arXiv:1409.1556}}
  (\bibinfo{year}{2014}).
\newblock


\bibitem[\protect\citeauthoryear{Wang, Yang, Xu, Hanjalic, and Shen}{Wang
  et~al\mbox{.}}{2017}]%
        {Wang-2017}
\bibfield{author}{\bibinfo{person}{Bokun Wang}, \bibinfo{person}{Yang Yang},
  \bibinfo{person}{Xing Xu}, \bibinfo{person}{Alan Hanjalic}, {and}
  \bibinfo{person}{Heng~Tao Shen}.} \bibinfo{year}{2017}\natexlab{}.
\newblock \showarticletitle{Adversarial cross-modal retrieval}. In
  \bibinfo{booktitle}{\emph{Proceedings of the 25th ACM international
  conference on Multimedia}}. \bibinfo{pages}{154--162}.
\newblock


\bibitem[\protect\citeauthoryear{Wang, Sahoo, Liu, Lim, and Hoi}{Wang
  et~al\mbox{.}}{2019}]%
        {Hao+CVPR2019_ACME}
\bibfield{author}{\bibinfo{person}{Hao Wang}, \bibinfo{person}{Doyen Sahoo},
  \bibinfo{person}{Chenghao Liu}, \bibinfo{person}{Ee-peng Lim}, {and}
  \bibinfo{person}{Steven~CH Hoi}.} \bibinfo{year}{2019}\natexlab{}.
\newblock \showarticletitle{Learning cross-modal embeddings with adversarial
  networks for cooking recipes and food images}. In
  \bibinfo{booktitle}{\emph{Proceedings of the IEEE Conference on Computer
  Vision and Pattern Recognition}}. \bibinfo{pages}{11572--11581}.
\newblock


\bibitem[\protect\citeauthoryear{Yan and Mikolajczyk}{Yan and
  Mikolajczyk}{2015}]%
        {Yan+CVPR2015}
\bibfield{author}{\bibinfo{person}{Fei Yan} {and} \bibinfo{person}{Krystian
  Mikolajczyk}.} \bibinfo{year}{2015}\natexlab{}.
\newblock \showarticletitle{Deep correlation for matching images and text}. In
  \bibinfo{booktitle}{\emph{Proceedings of the IEEE conference on computer
  vision and pattern recognition}}. \bibinfo{pages}{3441--3450}.
\newblock


\bibitem[\protect\citeauthoryear{Zagoruyko and Komodakis}{Zagoruyko and
  Komodakis}{2016}]%
        {wideResnet}
\bibfield{author}{\bibinfo{person}{Sergey Zagoruyko} {and}
  \bibinfo{person}{Nikos Komodakis}.} \bibinfo{year}{2016}\natexlab{}.
\newblock \showarticletitle{Wide residual networks}.
\newblock \bibinfo{journal}{\emph{arXiv preprint arXiv:1605.07146}}
  (\bibinfo{year}{2016}).
\newblock


\bibitem[\protect\citeauthoryear{Zhu, Ngo, Chen, and Hao}{Zhu
  et~al\mbox{.}}{2019}]%
        {zhu2019r2gan}
\bibfield{author}{\bibinfo{person}{Bin Zhu}, \bibinfo{person}{Chong-Wah Ngo},
  \bibinfo{person}{Jingjing Chen}, {and} \bibinfo{person}{Yanbin Hao}.}
  \bibinfo{year}{2019}\natexlab{}.
\newblock \showarticletitle{R2GAN: Cross-modal recipe retrieval with generative
  adversarial network}. In \bibinfo{booktitle}{\emph{Proceedings of the IEEE
  Conference on Computer Vision and Pattern Recognition}}.
  \bibinfo{pages}{11477--11486}.
\newblock


\end{thebibliography}

\end{document}